%% file: main.tex
\documentclass[a4paper,fleqn]{article}
\usepackage{INTERSPEECH2021,amssymb,amsmath,epsfig}
\usepackage{xspace}
\usepackage{srcltx}
\usepackage{caption}
\usepackage{subcaption}
\usepackage[export]{adjustbox}
\usepackage{multirow}
\usepackage{amsmath}
\usepackage{url}
\usepackage{hyperref}
\ninept
\def\reg{{\rm\ooalign{\hfil
     \raise.07ex\hbox{\scriptsize R}\hfil\crcr\mathhexbox20D}}}

\usepackage{color}
\usepackage[dvipsnames]{xcolor}

\newcommand{\CMT}[1]{{}}
\usepackage{soul}

\def\vx{\mathbf{x}}

\def\intent{\texttt{<intended>} }
\def\skip{\texttt{<unintended>} }

\def\reg{{\rm\ooalign{\hfil
     \raise.07ex\hbox{\scriptsize R}\hfil\crcr\mathhexbox20D}}}

\def\vx{\mathbf{x}}

\ninept
\begin{document}
\title{Streaming Intended Query Detection using E2E Modeling for Continued Conversation}
\name{Shuo-yiin Chang, Guru Prakash, Zelin Wu, Qiao Liang, Tara N. Sainath, Bo Li, Adam Stambler, 
Shyam Upadhyay, Manaal Faruqui, Trevor Strohman}
\address{Google Inc., U.S.A}
\email{\{shuoyiin,guruprakash,zelinwu,tsainath,boboli,wildstone,strohman\}@google.com}

\maketitle
\input{abstract}

\input{intro}

\input{model}
\input{experiments}
\input{results}

\input{conclusions}
\newpage
\bibliographystyle{IEEEbib}
\bibliography{main}
\end{document}

%% file: abstract.tex
\begin{abstract}
In voice-enabled applications, a predetermined hotword is usually used to activate a device in order to attend to the query. 
However, speaking queries followed by a hotword each time introduces a cognitive burden in continued conversations. 
To avoid repeating a hotword, we propose a streaming end-to-end (E2E) intended query detector that identifies the utterances directed towards the device and filters out other utterances not directed towards device. 
The proposed approach incorporates the intended query detector into the E2E model that already folds different components of the speech recognition pipeline into one neural network.
The E2E modeling on speech decoding and intended query detection also allows us to declare a quick intended query detection based on early partial recognition result, which is important to decrease latency and make the system responsive.
We demonstrate that the proposed E2E approach yields a 22\% relative improvement on equal error rate (EER) for the detection accuracy and 600 ms latency improvement compared with an independent  intended query detector. In our experiment, the proposed model detects whether the user is talking to the device with a 8.7\% EER within 1.4 seconds of median latency after user starts speaking. 
\\
\end{abstract}


%% file: intro.tex
\section{Introduction}
\label{sec:intro}

Streaming speech recognition systems have been successfully used in many voice interaction applications e.g. voice assistant and dialog systems. In a typical voice-enabled environment, a predetermined hotword e.g., "OK Google", "Alexa", "Hey Siri" is used to activate a device in order to attend to the primary query which follows the hotword \cite{Huang2019, Tang2018, Wollmer2019}. While hotword activation is commonly used for single query scenario, having to speak the hotword each time disrupts the flow of continued conversation. For example, the user may speak "OK Google, what’s the weather like today?” After the system answers, the user has to speak “OK Google, what about tomorrow?” to get the forecast. This repeated hotword phrase can cause a cognitive burden on the user.

To get rid of the hotword on follow-on queries, it is crucial to build a classifier that identifies the utterances directed towards the device, referred to as intended-queries, while suppressing non-device directed utterances, referred to as unintended-queries e.g. speech to other individuals, thinking aloud or random speech. Here, we term this classifier as the intended-query (IQ) detector. 
%\BL{maybe explain a bit we are not removing the 1st hotword? theorectically, with IQ we can totally get rid of hotwords.

\begin{figure}[h!]
\centering
\vspace{-0.1in}
\hspace{-0.05in}
    \includegraphics[scale=0.28]{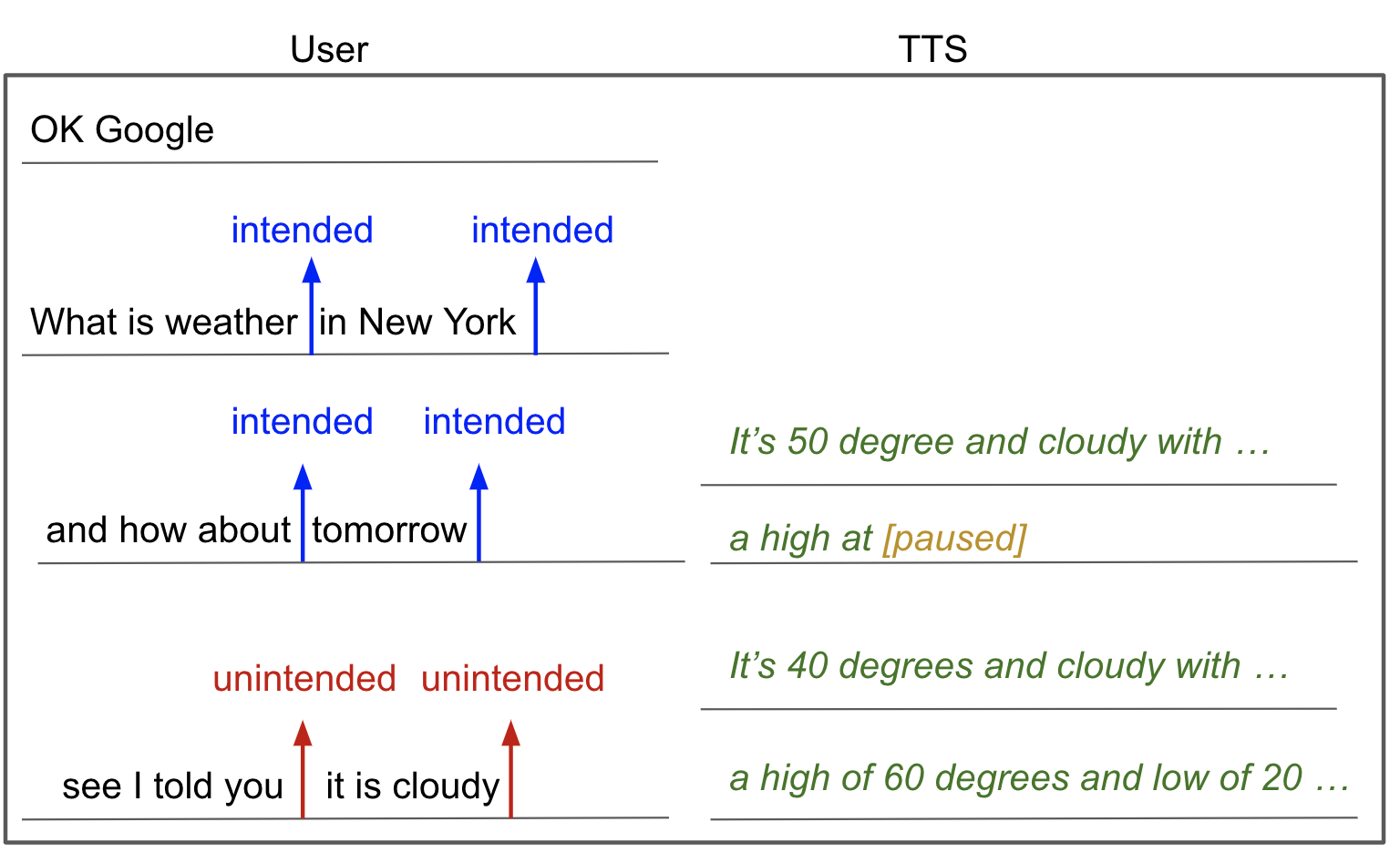}
    \vspace{-0.1in}
    \caption{Intended-query detection for follow-on queries}
    \vspace{-0.05in}
    \label{fig:intro_example}
\end{figure}

Figure \ref{fig:intro_example} demonstrates the scenario of continued conversation. The system stays open for any follow-on utterances after a user speaks the first query using the hotword. Subsequent queries are then filtered by intended-query detector that makes a series of binary decisions: whether to barge-in the system and send recognition results downstream for intended-queries or to discard for unintended-queries \cite{patent20}.
It is desirable to identify the intended-queries as soon as possible to respond to the user quickly while avoid accepting unintended-queries.
%which could cause privacy concern and redundant downstream computation. As privacy is a highly concerning issue, Google already works hard to eliminate unintended speech through many system components in addition to speech interface, thus, the data is not logged if it's considered not for Google.
Accepting unintended-queries would generate incorrect responses to the user, resulting in a very jarring experience. On the other hand, rejecting intended-queries or slow detection makes the system feel unresponsive.

%\section{Overview and Related Works: WIP}
%\label{sec:past_work}

Recent studies on intended-query detection have investigated acoustic-based approaches \cite{AM19, AM13} e.g. prosody, or using ASR decoder features \cite{lex13, amazon18} e.g. lexical features or confidence. For example, in \cite{patent20, amazon18, amazon19}, word or subword embeddings from the top ASR hypothesis are explored as extra features in addition to acoustic and decoder features. The features from different sources are typically encoded by separate models and concatenated to a single feature vector per utterance. The utterance features are then fed through a fully connected network to compute the probability of the intended query. However, the utterance-level classifiers may need to consume a complete utterance to declare an IQ decision, which is undesirable for many streaming applications. For example, the system is expected to pause immediately if a user speaks to the device in the middle of system response as given by the example in Figure \ref{fig:intro_example}.
%The previous works in \cite{patent20} build an an intended-query detector incorporating acoustic and semantic models. 

%\TS{were the previous papers streaming for IQ detection? If not, you should point this out.}
In contrast in this paper, we propose to build a sequence-to-sequence model that incorporates ASR and intended-query detection. This allows us to not only do ASR, jointly with intended-query detection, but also allows us to do the detection in a streaming fashion, which is important for latency sensitive applications.

E2E models look at folding different components of the speech recognition pipeline (i.e., the acoustic, pronunciation, and language models) into one neural network \cite{li2020comparison,Ryan19,CC18,KimHoriWatanabe17,JinyuLi2019,Zeyer2020}. Recent works have shown that folding further tasks into the model, for example endpointing, results in better WER and latency tradeoff, compared to having separate tasks \cite{Shuoyiin19,li2020towards}. In the spirit of joint optimization, in this paper, we proposed to build a joint ASR decoding and IQ detection with an E2E RNN-T model. To achieve this goal, we look to train the RNN-T model with additional tokens. Specifically, we incorporate the special symbols \intent which indicates the intended queries and \skip for unintended queries as part of the expected label sequence e.g. "\textit{snooze alarm} \intent \textit{at 8:00}". 
The RNN-T model then makes IQ decisions based on the IQ tokens. Thus, the model acts jointly as a decoder and an IQ detector in a streaming way. 

%In many streaming conversational voice interaction applications, it is crucial to react to the intended query as quickly as possible even before the system has finished. For example, the system is expected to pause the TTS immediately if a user speaks to the device in the middle of system response as given by the example in Figure \ref{fig:intro_example}. The joint optimization with a streaming RNN-T can achieve the low latency detection by learning to declare the IQ tokens with early recognition results.  
%\TS{not clear what you mean by early partial here}. 
%\TS{i think you can remove this sentence here since its too much detail from the intro:} In this work, we insert the label sequences with the IQ tokens based on the slot filling used in the natural language understanding (NLU) model \cite{Googleblog20} and the silence alignment.

%TS{i think it will help to add some sentences on why latency is important and how doing this joint optimization with a streaming RNN-T system helps to achieve this goal. this also differentiates it better from previous work.} %This allows us to fold intended-query detector jointly into the E2E model, minimizing the dependence on an external detector. 

We investigate the detection error tradeoff (DET) graph i.e. false accept and false reject rate to evaluate the quality of IQ detector. To evaluate the detection latency, we compute the latency, defined as the time between users starting speaking and the system declaring the IQ decision. The proposed E2E model achieves an 8.7\% EER and a median latency of 1.4 seconds, which yields 22\% and 600 ms improvement relative to the independent baseline detector. %\mfar{i wonder if its worth metioning that predicting the extra ``intended'' or ``unintended'' token every few steps would reduce the speed of decoding}

%To fully leverage the encoder and prediction networks from the RNN-T ASR model, we introduce another joint layer, referrer as intended-query detection layer on top of encoder and prediction network and emit two special tokens \intent and \skip indicating intended and unintended queries respectively. By sharing the parameters from encoders and prediction network of RNN-T, we can exploit acoustic representations and intonation patterns from encoder and semantically intended sentences by decoder. 

%The recent development of end-to-end (E2E) models \cite{li2020comparison,Ryan19,CC18,JinyuLi2019,Zeyer2020}  has largely improved ASR performance. There have been many successful efforts in build a streaming end-to-end ASR system with better quality and latency \cite{sainath20streaming, bo21system, sainath2021efficient} and has been shown to outperform a strong conventional-based ASR model\cite{Golan16}. 

%As the consequence of success in E2E models, we proposed to build the intended-query model based on the on-device E2E system. To fully leverage the encoder and prediction networks from the RNN-T ASR model, we introduce another joint layer, referrer as intended-query detection layer on top of encoder and prediction network and emit two special tokens \intent and \skip indicating intended and unintended queries respectively. By sharing the parameters from encoders and prediction network of RNN-T, we can exploit acoustic representations and intonation patterns from encoder and semantically intended sentences by decoder. 

%% file: model.tex
\section{Method}
\label{sec:model}

\subsection{Baseline Systems}   \label{sec:base_model}
We investigate two external IQ detectors as baseline systems to compare with the proposed E2E IQ detector. 

\subsubsection{Acoustic IQ Detector}  \label{sec:am_model}
Figure \ref{fig:system_base}a depicts the IQ detector based on acoustic features, referred as acoustic IQ detector.
The acoustic IQ detector consists of long short term memory (LSTM) layers, which take  typical acoustic features (i.e. log-mel filterbanks) as input and optimize for the frame-level targets of the two classes, intended or unintended. To obtain the frame-level targets, we simply label each frame of the entire utterances from Google’s assistant traffic as intended while label the frames of utterances from Google’s other speech applications e.g. YouTube as unintended.  
Thus, the acoustic IQ detector can extract distinctive acoustic patterns e.g. speech tempo to determine whether the audience for the utterance is likely talking to the device.  %\TS{how do you get the labels for the data, you might want to clarify like you did in the e2e section.}
To declare the final IQ decisions, we first obtain the frame-level decisions by thresholding the posterior of the intended class. Next, a state machine is applied on top of the frame-level decisions to smooth the distribution.
%The acoustic model consists of long short term memory (LSTM) layers to encode the typical acoustic features (i.e. log filterbank) and computes an output of a fixed-size acoustic embedding per utterance.
%Also, it reduces the failures due to misrecognitions.
\subsubsection{IQ Detector using Acoustic and Text embeddings}  \label{sec:am_lm_model}
\begin{figure}[h!]
\centering
\vspace{-0.15in}
\hspace{-0.05in}
    \includegraphics[scale=0.31]{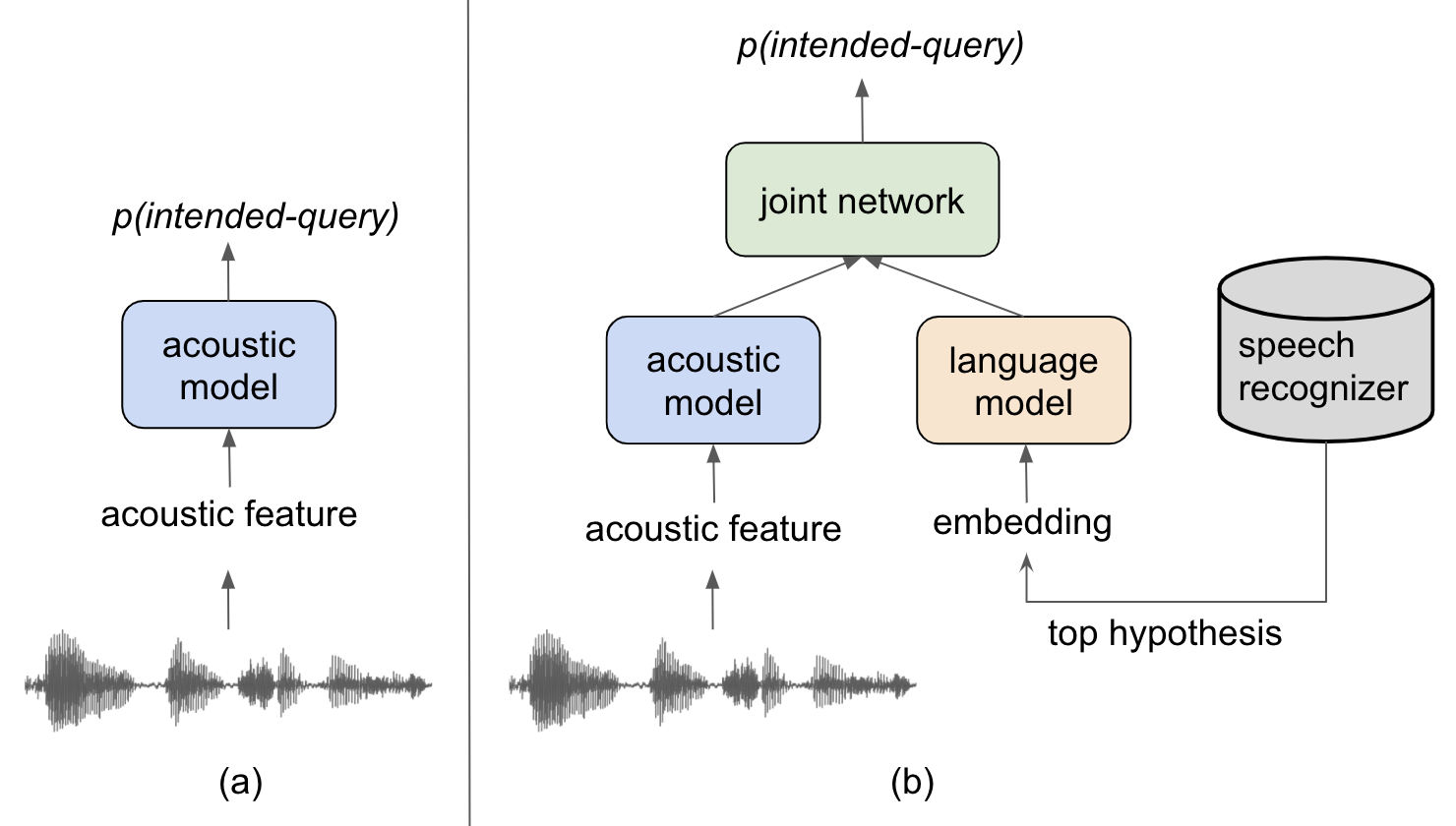}
    \vspace{-0.1in}
    \caption{Baseline systems: (a) acoustic IQ detector and (b) acoustic-text IQ detector}
    \vspace{-0.15in}
    \label{fig:system_base}
\end{figure}

In addition to acoustic patterns, the spoken words also provide useful cues for differentiating between intended and unintended queries. Hence, a language model that directly models word sequences could improve the IQ detector. 
Figure \ref{fig:system_base}b illustrates the IQ detector incorporating both the acoustic and word sequences following \cite{patent20}, termed as acoustic-text IQ detector.
%The system consists of 3 components, acoustic model, semantic model and joint network.  
%The acoustic model consists of long short term memory (LSTM) layers to encode the typical acoustic features (i.e. log filterbank) and computes an output of a fixed-size acoustic embedding per utterance.
%The system leverages the acoustic model to extract distinctive acoustic patterns e.g. speech tempo and accent to assist in determining whether the audience for the utterance is likely talking to the device. Also, it reduces the failures due to misrecognitions.

To encode acoustic sequences, we take the hidden states of the last frame from the LSTM acoustic model to obtain a fixed length acoustic embedding per utterance.
%\TS{you might want to mention this is not streaming now}.  
On the other hand, the language model encodes word sequences from top hypothesis of the speech recognizer. The hypothesis is first converted to a word embedding and then encoded with a convolutional neural network (CNN). 
The CNN uses max pooling to decrease the input and reduce the computational complexity of the network. 
%Max pooling allows for significant reduction in data through the filtering and averaging the input without degrading the quality. 
Once the acoustic and text embeddings have been extracted, the embeddings are then concatenated and fed through a fully connected joint network to predict the probability of intended query and unintended query. 

The model incorporate both acoustic clues and text patterns to provide a more accurate detection accuracy compared with the IQ detector based on pure acoustic features. However, the acoustic and word sequences are encoded to utterance-level embeddings, thus the model could learn to predict the IQ decisions until it has seen the embeddings of a complete utterance when used in a streaming system. To address the problem, we propose to build a streaming E2E IQ detector built on ASR network in the following sections. 

%\mfar{so the baseline system computes a single prediction for the entire utterance, but the proposed model computes it at multiple timesteps. how do we compare them, maybe this will be answered later.}
%One embedding is output for each feature vector. The acoustic encoder is an independently trained frame-level classifier where the frame-level targets are obtained by repeating the utterance label.  

\subsection{E2E Model Labeling using Slot Filling}   \label{sec:nlu}

To build a sequence-to-sequence E2E ASR and IQ model, we insert \intent to the label sequences of intended queries collected from Google assistant while \skip to unintended queries obtained from YouTube. A straightforward question with E2E ASR models that predict subword units (i.e., wordpieces), is how to insert special symbols i.e. IQ tokens, into the original label sequence. 
Our goal is to be able to detect \intent as quickly as possible so the system and downstream NLU components can perform the appropriate action. In one extreme, if \intent is inserted as the first symbol, while this allows a quick intent decision, the decision is unreliable as it is based only silence patterns before speaking and empty semantic history in streaming systems. In another extreme, if the symbol is appended at the utterance end, it could only emit the \intent once finishing decoding the whole utterance, which is slow. Thus, our goal is to insert \intent on incomplete transcriptions where a true intent has been initiated. 

\begin{figure}[h!]
\centering
\hspace{-0.05in}
    \includegraphics[scale=0.38]{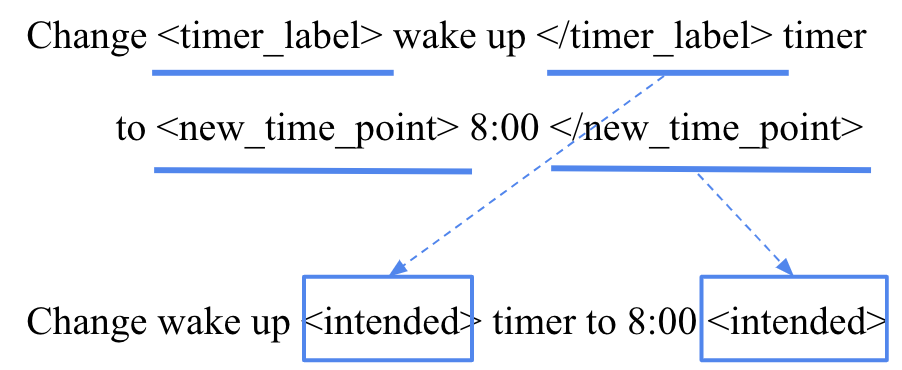}
    \vspace{-0.1in}
    \caption{Labeling using slot filling}
    \vspace{-0.15in}
    \label{fig:slot}
\end{figure}

To achieve this goal, we exploit using slot filling and silence alignment as indicators of where to insert the IQ tokens. First, we extract the slots by parsing the transcriptions through an NLU model \cite{Googleblog20, ICLR17} 
%\mfar{let's skip saying this is used for assistant -- NLU team doesn't reveal the models being used to public} 
to obtain the span of the slots as shown in Fig. \ref{fig:slot}. Each slot represents a common semantic concept in the Google's voice assistant traffic e.g. \texttt{<media\_object>}, \texttt{<time\_label>}, etc. Thus, the slot tags identify the parts of the transcriptions involving a complete semantic content. We then convert the closing tag of the slot into the \intent tokens as Fig. \ref{fig:slot}. In addition, we also insert the \intent or \skip at each silence segment to further force emitting IQ decisions quicker. These silence segments, indicating the user is pausing, are identified based on forced alignment. 

%\begin{center}
%Label with Slot:\\
    %\quad\texttt{Change <TimerLabel> wake up </TimerLabel> timer to} \\
    %\quad\texttt{Play<MediaObject>Paramore</MediaObject>}\\
    %\quad\texttt{on <App> Spotify </App>}\\
    %\quad\texttt{<NewTimePoint> 8:00 </NewTimePoint>} \\
%\\
%Label:\\
%    \quad\texttt{Play Paramore <intended>}\\
%    \quad\texttt{on Spotify <intended>}\\
    %\quad\texttt{Change wake up <intended> timer to 8:00 <intended>} \\
%\end{center}

\subsection{RNN-T ASR and IQ detector}   \label{sec:arch}

\begin{figure}[h!]
\centering
\hspace{-0.05in}
    \includegraphics[scale=0.3]{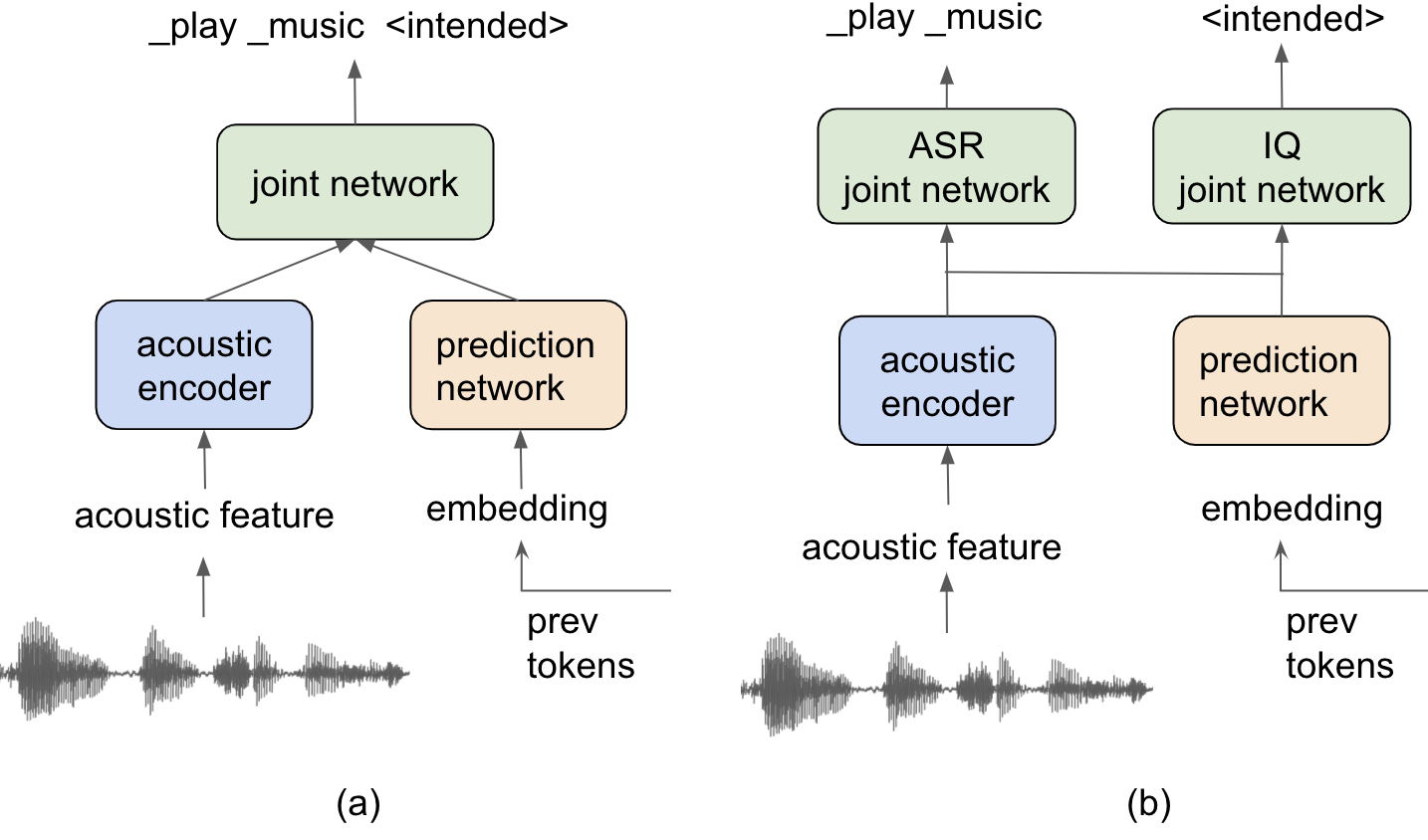}
    \vspace{-0.1in}
    \caption{E2E model architectures using (a) convectional RNN-T and (b) RNN-T with additional joint network}
    \vspace{-0.15in}
    \label{fig:system_e2e}
\end{figure}

As illustrated in Fig. \ref{fig:system_e2e}a, we use the RNN-T architecture while augmenting the label sequence with the IQ tokens as described in Sec. \ref{sec:nlu} to build the E2E model. 
Thus, at each time step $t$, the model receives a new acoustic frame $x_t$ and outputs a probability distribution over $y_t \subset \{V \cup \intent \cup \skip\}$, V being the wordpiece vocabulary and a blank symbol.

%\TS{it's a language model not a semantic model}
The RNN-T consists of 3 components: encoder, prediction network and joint network, which is analogous to acoustic model, language model and joint network in the baseline acoustic-text IQ detector  (Fig. \ref{fig:system_base}b) as described in Sec. \ref{sec:am_lm_model}. The encoder acts as the acoustic model. At each time step, the encoder receives typical acoustic features and outputs acoustic encodings. 
The recurrent prediction network obtains the embedding vectors from $N$ previous (non-blank) labels \cite{Rami21} and produces the text encodings. Hence, the prediction network works as the language model of the acoustic-text IQ detector.
The encoder and prediction network are then fed into the joint network that projects acoustic and text encodings to a fully-connected layer to computes the output logits.

However, simply introducing IQ tokens into the standard RNN-T model could degrade ASR quality. While the wordpiece history is useful for IQ prediction, the IQ tokens do not provide informative features for wordpieces prediction.  Instead, including IQ tokens could decrease the context window as the prediction network attends to only a limited context \cite{Rami21}. In our experiments, we observed 4\% relative WER increase. 

To ensure that recognition quality is consistent with the standard E2E model, we adapt the model architecture by introducing separate joint networks, referred as ASR joint network and IQ joint network as illustrated in Fig. \ref{fig:system_e2e}b. The ASR joint network provides the probability distribution over the wordpieces space as the conventional ASR while the IQ joint network is responsible for IQ decisions.

%\TS{discuss the 2nd joint a bit better, and cite previous work. also motivate why you want to do 2nd joint (so you dont degrade the 1st-hypothesis by inserting extra symbols.}
%To ensure that recognition quality is consistent with the conventional ASR, we introduce a separate joint layer for intended-query detection as illustrated in Figure~\ref{fig:system_e2e}.
%The IQ layer was built on top of encoders and prediction networks, hence, it could leverage both acoustic pattern and semantic history for the intended-query detection. The IQ layer is responsible for the prediction of \intent or \skip while the original joint layer remains unchanged. 

In training phrase, we perform two stages training strategy. First, we optimize the encoder, prediction network and the ASR joint network using the regular wordpieces label sequences for recognition quality. Next, we freeze the encoder and prediction network while initialize the IQ joint network with the ASR joint network. The IQ joint network is then fine-tuned with the expanded label sequence including wordpieces and IQ tokens to adapt the parameters with respect to additional loss due to IQ tokens insertion. Thus, the IQ joint network is able to predict distributions of IQ tokens given the existing acoustic and wordpieces label encodings. During inference, we rely on only the ASR joint network for beam search decoding over the wordpiece space:
\begin{equation}
y* = \operatorname*{arg\,max}_y \log P_{asr}(y|\vx_{t-k}, \cdots, \vx_{t}, y_{u-N}, \ldots, y_{u})
	\label{eqn:beam-search}
\end{equation}
where $\mathbf{x}_t,... \mathbf{x}_{t-k}$ represents acoustic observations received by the Conformer encoder \cite{gulati2020conformer} with a context window of \textit{k} and $\mathbf{y}_{u-N},... \mathbf{y}_u$ stands for the wordpiece sequences. At each time step, the IQ joint network computes the the probability of the IQ tokens given the decoding paths obtained by ASR joint network using Eq. \ref{eqn:beam-search}. The posterior of IQ tokens can be expressed as: 
\begin{equation}
    \begin{aligned}
	P_{iq}(\intent|\vx_{t-k}, \cdots, \vx_{t}, y_{u-N}, \ldots, y_u), \\
	P_{iq}(\skip|\vx_{t-k}, \cdots, \vx_{t}, y_{u-N}, \ldots, y_u)
	\end{aligned}
	\label{eqn:joint}
\end{equation}
Thus, the system could make a series of IQ decisions at each time step by thresholding the posterior of \intent obtained by Eq. \ref{eqn:joint}. In order to emit the tokens early, we use the FastEmit \cite{yu21fastemit} loss to encourage paths that outputs tokens earlier.

%% file: experiments.tex
\section{Experiments \label{sec:experiments}}

\subsection{Datasets}

The training data covers utterances of intended voice queries and unintended (nondevice-directed) voice transcriptions. The intended voice query data consists of around 15 million utterances collected from Google's voice search and actions on various mobile platforms and Google Home. The unintended voice transcription includes around 10 million of utterances obtained from YouTube.  The utterances are anonymized andhand-transcribed. In addition to the diverse training sets, multi-condition training (MTR) \cite{Chanwoo17} are also used to further increase data diversity.

The evaluation set includes 53k utterances of proxy intended queries and 18k utterances of proxy unintended queries. The proxy voice queries are collected from 300 speakers. The proxy unintended queries covers 25 specific domains to include the scenarios of talking to other individuals, thinking aloud or random speech. Table \ref{table:neg_prompts} demonstrates the prompts for some proxy unintended query domains. The proxy intended queries include 18 domains with common voice actions, questions, search queries and queries in dialog scenarios.

\begin{table} [h!]
\vspace{-0.05in}
\caption{prompts of proxy unintended query.}
\centering
\vspace{-0.1in}
\begin{tabular}{|c|} \hline
	Prompts of proxy unintended queries \\ \hline \hline
	Speak to your spouse about food/TV series or politics \\ \hline
	Speak to your spouse when Assistant helps you well \\ \hline
	Express you love the song (to yourself) Assistant played \\ \hline
%	Repeat a line from a movie or a song \\ \hline
	Speak something that pops to you but on a very low voice  \\ \hline
\end{tabular}
\vspace{-0.15in}
\label{table:neg_prompts}
\end{table}

\subsection{Model Description}
The acoustic features used for the models are 128D log-mel filterbanks. 
%computed using 32 msec windows with a 10 msec frame step. 
%Features from 4 contiguous frames are stacked to form a 512 dimensional input representation, which is further subsampled by a factor of 3 and passed to the model. 
%SpecAugment \cite{Park2019} is used to improve models’ robustness  against  noise  where  two  frequency  masks  with  a  maximum length of 27 and two time masks with a maximum length of 50 are used.
Following \cite{gulati2020conformer}, the RNN-T encoder network consists of 12 Conformer layers where each layer is of 512 dimension with a time-reduction layer after the 2nd layer.
The Conformer layers consist of causal convolution and left-context attention layers where 8-head attention is used in the self-attention layer and the convolution kernel size used is 15. 
The embedding prediction network \cite{Rami21} uses an embedding dimension of 320, and has 1.96M parameters.
The ASR joint network and IQ joint network consists of a single feed-forward layer with 640 units.
The RNN-T are trained with HAT factorization \cite{Biadsy20} to predict 4,096 word pieces \cite{Schuster2012} and two IQ tokens. The RNN-T model has 140M parameters. 
It is optimized by minimizing the RNN-T loss. The FastEmit \cite{yu21fastemit} regularization with a weight of 5e-3 is explored to improve the model's prediction latency. We don't use 2nd-pass model for the experiments here.

The baseline acoustic IQ detector used 3 LSTM layers with 64 units.
The acoustic-text IQ detector follows the architecture in \cite{patent20}.
The language model combines the CNN word embedding layer with a 2 layer LSTM run over the utterances while the acoustic model consists of 3 LSTM layers with 64 units.
The 100 units from the word embedding CNN’s max pooling layer are concatenated with the 64 units of the hidden state from the 2nd layer of the LSTM. 
These 164 units are then fed into two fully connected layers.

%All models are trained in Tensorflow \cite{AbadiAgarwalBarhamEtAl15} using the Lingvo \cite{shen2019lingvo} toolkit on 8 × 8 Tensor Processing. Models are optimized using synchronized stochastic gradient descent. We use the Adam optimizer \cite{KingmaBa15} with parameters $\beta_1$=0.9 and $\beta_2$=0.999.

%The baseline system is an external intended-query classifier consisting of an acoustic encoder and a text encoder. The acoustic encoder consists of X layers of LSTMs receiving acoustic features as input while the the CNN word embedding layers encode the top hypothesis. The joint network consisting of X layers of LSTMs then output the distribution of intended-query and unintended-queries. The proposed architecture share the acoustic encoder and word embedding that has been used by RNN-T speech recognizer while use a separate joint network is for intended-query and unintended-query. The label sequence consists of the transcriptions as well as the \intent and \skip inserted based on the NLU slot.  

%% file: results.tex
\section{Experiments \label{sec:results}}

\subsection{Evaluation metrics}

The DET curves are frequently used to describe a binary classification task. 
Here, we report the DET curve (false rejection against false accept) for the intended-query detection. Lower curves are better. 
The posterior of the intended-query is thresholded to obtain the IQ decision. 
A false accept would introduce incorrect responses to the user and extra computations for downstream components. A false rejection means the system is unresponsive. 
We use the EER to evaluate the performance of each approach. 
We also calculate the 50th latency and 90th latency of IQ detection where the latency is measured between the timestamp of start-of-speech and the timestamp of the intended query decision made. The latency is only calculated for the test sets of intended queries as the latency of declaring intended query is useful for quicker response.

\subsection{Results}
\begin{figure}[h!]
\centering
\hspace{-0.05in}
    \includegraphics[scale=0.45]{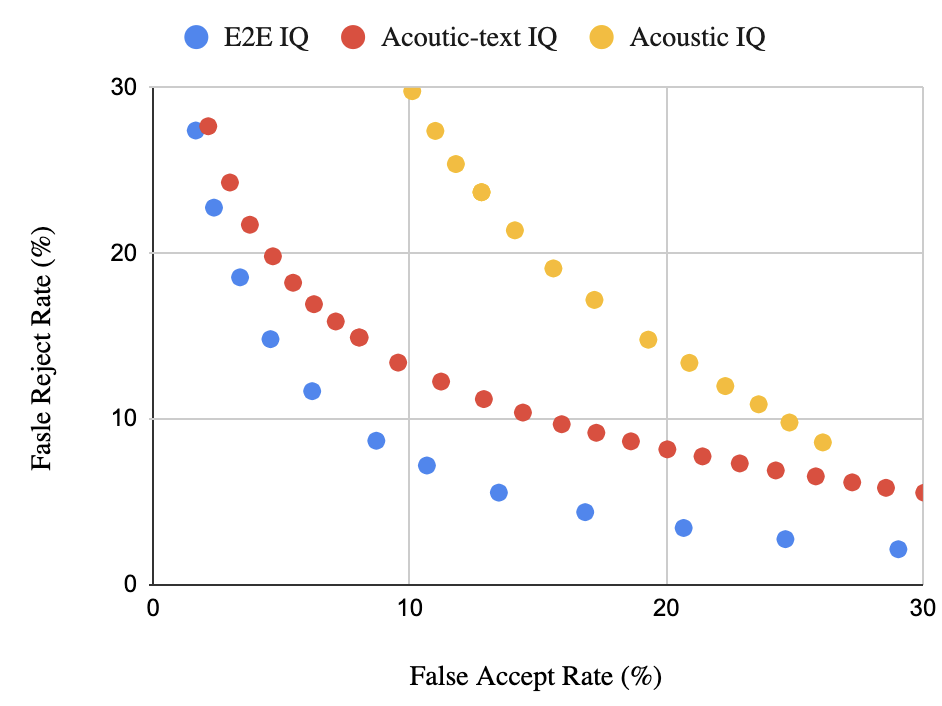}
    \vspace{-0.1in}
    \caption{DET for acoustic IQ detector, acoustic-text IQ detector and E2E IQ detector}
    %\vspace{-0.1in}
    \label{fig:frfa}
\end{figure}

Table \ref{table:eer_latency} demonstrates the EER of the baseline acoustic, acoustic-text IQ detector and the proposed E2E IQ detector. As shown in Table \ref{table:eer_latency}, the EER of the E2E IQ detector provides 22\% relative improvement compared to the acoustic-text detector and is 49\% better compared to the acoustic detector. 
As shown in Figure \ref{fig:frfa}, the E2E IQ detector provides a consistently better FR/FA tradeoff compared to the two baseline models, especially for the operations points of FR below 10\%,  both of which degrade FA rapidly for the region. 
%Table demonstrates the Equal Error Rate of the E2E approach and the baseline approach. 
%The acoustic IQ detector increased rapidly in the region of 

For the latency metrics in Table \ref{table:eer_latency}, we observe the acoustic IQ detector provides the lowest latency to declare the IQ decisions despite higher EER compared with the other approaches. The acoustic-text detector has the highest 50th and 90th percentile latency. This implies that early incomplete audios and hypotheses often fail to trigger the IQ decisions where the long-tail examples require around 2.8 seconds to identify that users are talking to the device. Comparing to the acoustic-text detector, the proposed streaming E2E IQ detector could speed up the 90th percentile latency by 600 ms. The latency evaluation show that E2E IQ detector is only 60 ms slower to acoustic IQ for the 90th percentile latency while provides a much better EER. 
%Thus, the acoustic-text IQ detector using the embeddings extracted over the entire utterances during training needs more audio signals to declare a decision when operated in a streaming system.
%For the latency metrics, the acoustic IQ detector performs the quickest decisions as it's a frame-level classifier. 

\begin{figure}[h!]
\centering
\hspace{-0.05in}
    \includegraphics[scale=0.38]{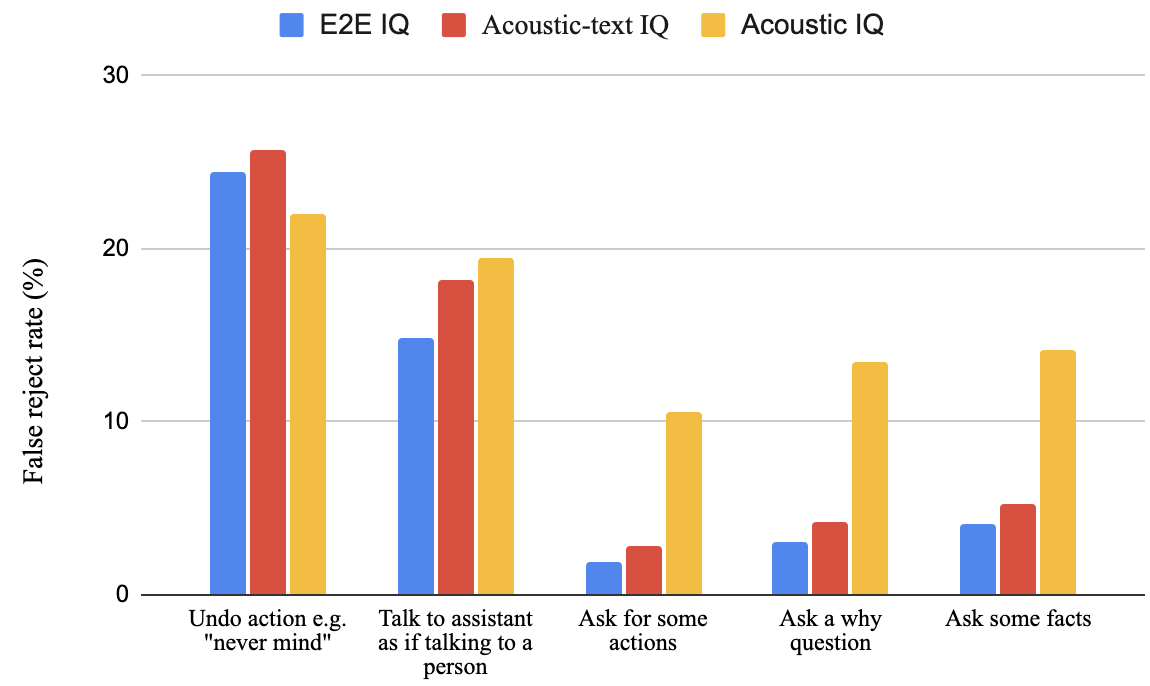}
    \vspace{-0.1in}
    \caption{FRs of specific domains}
    %\vspace{-0.15in}
    \label{fig:example_fr}
\end{figure}

Figure \ref{fig:example_fr} reports the FR rates of each model for 5 selected domains of intended queries to investigate the errors. We evaluate the models with the operation points at EERs. The E2E IQ is better than both baseline systems for common actions e.g. music related actions, making a call, creating events, etc. Similar trends could be observed for the queries about asking facts e.g. "how many metres in a mile" or questions "why is the sky blue". The E2E IQ shows consistent gains over the other systems except for a specific domain (undo action) that has fewer training examples where the acoustic IQ is more robust.

\begin{table} [h!]
%\vspace{-0.05in}
\caption{EER and latency.}
%vspace{-0.1in}
\centering
\begin{tabular}{|c|c|c|c|c} \hline
    model &  EER  & 50th & 90th  \\
	      &    (\%)  & latency & latency  \\ \hline
	Acoustic IQ & 17.2 & 1200 ms & 2160 ms\\ \hline
	Acoustic-Text IQ & 11.2 & 1860 ms & 2820 ms\\ \hline
	E2E IQ &  8.7 & 1440 ms & 2220 ms\\ \hline
\end{tabular}
\label{table:eer_latency}
\vspace{-0.15in}
\end{table}

%% file: conclusions.tex
\section{Conclusions \label{sec:conclusions}}
In this work, we a propose a sequence-to-sequence model that incorporates ASR and intended-query detection into a unified E2E RNN-Transducer.
By joint modeling, the E2E model allow us to do intended-query detection and ASR jointly in a streaming fashion to achieve a fast and accurate intended-query detection.
By comparing to a separate system, results and analyses show that the E2E model reduces the ERR from 11.2\% to 8.7\% and provides 600ms latency improvement.
%\section{Acknowledgement \label{sec:ack}}